\documentclass[runningheads]{llncs}
\usepackage[T1]{fontenc}
\usepackage{graphicx}
\usepackage{hyperref}
\usepackage{amsmath}
\usepackage{amssymb}
\usepackage[numbers]{natbib}
\usepackage{algorithm}%
\usepackage{algorithmicx}%
\usepackage{algpseudocode}%

\usepackage{orcidlink}
\usepackage[misc]{ifsym}
\begin{document}
\title{GFlowNets for Learning Better Drug-Drug Interaction Representations}
% \author{Anonymous submissionr\inst{1}\orcidID{0000-1111-2222-3333} \and
\author{Azmine Toushik Wasi\inst{({\text{\Letter}})}\orcidlink{0000-0001-9509-5804}}
\authorrunning{A. T. Wasi et al.}
\institute{Shahjalal University of Science and Technology, Bangladesh\\
\email{azmine32@student.sust.edu}}
\maketitle              % typeset the header of the contribution
\begin{abstract}
Drug–drug interactions pose a significant challenge in clinical pharmacology, with severe class imbalance among interaction types limiting the effectiveness of predictive models. Common interactions dominate datasets, while rare but critical interactions remain underrepresented, leading to poor model performance on infrequent cases. Existing methods often treat DDI prediction as a binary problem, ignoring class-specific nuances and exacerbating bias toward frequent interactions. To address this, we propose a framework combining Generative Flow Networks (GFlowNet) with Variational Graph Autoencoders (VGAE) to generate synthetic samples for rare classes, improving model balance and generate effective and novel DDI pairs. Our approach enhances predictive performance across interaction types, ensuring better clinical reliability.

\end{abstract}

\section{Introduction}
Drug–drug interactions (DDIs) represent a critical issue in clinical pharmacology, as adverse interactions can lead to significant patient harm and reduced therapeutic efficacy \cite{10.1093/bib/bbad445}. Numerous computational models have been developed to predict DDIs, leveraging diverse features from chemical structures to biological networks \cite{10.1093/bib/bbaa256,sgaessgaweg34asg}. However, a pervasive challenge in this domain is the severe class imbalance among interaction types. Common interaction types, such as synergistic effects or well-characterized adverse reactions, dominate the datasets, while rare interaction types remain under-represented \cite{Ezzat2016}. This imbalance hinders the model’s ability to learn nuanced patterns associated with infrequent interactions. As a consequence, predictive performance on rare, yet clinically significant, interaction types suffers. The disparity in data distribution thus poses an urgent need for innovative solutions in DDI prediction.

In light of the class imbalance, existing state-of-the-art methods are often trained in a binary setting \cite{10.1093/bib/bbaa256,wasi2024cadglcontextawaredeepgraph,ngo2022predictingdrugdruginteractionsusing}, treating the DDI prediction task as a simple presence-or-absence problem. This binary framing tends to disregard the inherent heterogeneity among different interaction types. Consequently, the models become biased towards common interaction types and fail to adequately capture the underlying characteristics of rare classes. The motivation for this work stems from the recognition that a one-size-fits-all approach is insufficient for capturing the diversity of DDIs. Addressing the imbalance is crucial for improving the reliability and clinical utility of these predictions \cite{10.1093/bib/bbaa256}. By acknowledging and explicitly targeting the disparity in interaction frequencies, we aim to bridge the gap between theoretical performance and real-world application with GFlowNets \cite{nica2022evaluating,roy2023goalconditionedgflownetscontrollablemultiobjective}.

To mitigate the challenges posed by class imbalance, we propose an innovative framework that integrates a Generative Flow Network (GFlowNet) \cite{JMLR:v24:22-0364,Jain_2023} module with a Variational Graph Autoencoder (VGAE) \cite{ngo2022predictingdrugdruginteractionsusing, wasi2024cadglcontextawaredeepgraph}. Our approach first computes a reward for each interaction type that is inversely proportional to its frequency, thereby guiding the sampling process towards under-represented classes. The GFlowNet module sequentially generates synthetic DDI samples by first selecting an interaction type based on this reward and then sampling a drug pair conditioned on that type. These synthetic samples are then used to augment the original training data, effectively balancing the class distribution. Experimental results indicate that this method enhances the model's ability to predict both common and rare interaction types with improved robustness. The proposed framework not only addresses a critical limitation in current DDI prediction models but also holds promise for broader applications in imbalanced classification problems across biomedical domains.

\section{Related Works} \label{RelatedWorks}
Drug discovery tasks like drug combination and drug-drug interaction (DDI) prediction have been widely studied. DDI prediction, the focus of this work, aims to identify synergistic, antagonistic, or neutral interactions between drugs. Recent advances leverage graph-based architectures like MFConv \cite{MFConv} and GraphDTA \cite{10.1093/bioinformatics/btaa921} for molecular representation. Lim et al. \cite{Lim2019PredictingDI} integrate 3D structures and graph attention, while Torng and Altman \cite{Torng2018GraphCN} employ graph autoencoders for protein-pocket representations. Autoencoders such as VGAE \cite{VGAE-main} and GraphVAE \cite{Simonovsky2018GraphVAETG} generate latent graph embeddings, aiding in molecular structure modeling. GraphMAE \cite{Hou2022GraphMAESM} introduces masked pretraining for robust graph learning. Knowledge-based autoencoders like \cite{10.1093/bib/bbaa256} incorporate semantic embeddings for DDI tasks. Self-supervised learning methods such as SupDTI \cite{Chen2022PredictingDI} and SuperGAT \cite{kim2021how} improve node representations using unlabeled data. Contrastive frameworks like HeCo \cite{Wang2021SelfsupervisedHG} and SHGP \cite{7802e5f1421cdc61595fb529fde9dc8b57abf489} enhance graph pretraining via meta-paths and structure-aware pseudo-labeling. Our work builds on these ideas by proposing a VAE-based DDI prediction model enriched with graph-aware self-supervised objectives.

Unlike prior models that treat DDI prediction as a binary classification task \cite{10.1093/bib/bbaa256, ngo2022predictingdrugdruginteractionsusing, wasi2024cadglcontextawaredeepgraph}, our work addresses the overlooked issue of class imbalance across diverse interaction types. Existing approaches often ignore the semantic heterogeneity of DDIs, leading to poor performance on rare but clinically significant classes. We propose a novel integration of GFlowNet and VGAE to explicitly model interaction-type distributions, where GFlowNet generates balanced synthetic samples based on reward-guided sampling. This framework not only improves prediction robustness across all classes but also provides a generalizable strategy for imbalanced biomedical graph problems.

\section{Model Architecture}
Our framework tackles the class imbalance in Drug-Drug Interaction (DDI) prediction by synergizing a VGAE with a GFlowNet. The VGAE first learns a rich, graph-based latent representation of drugs. The GFlowNet is then trained to generate synthetic DDI samples, focusing on rare yet plausible interactions. These generated samples augment the original dataset, leading to a more robust and balanced training process for the final prediction model.

\subsection{Variational Graph Autoencoder (VGAE) for DDI Representation Learning}
We model the set of known DDIs as a multi-relational graph $\mathcal{G} = (\mathcal{D}, \mathcal{E})$, where $\mathcal{D}$ is the set of drug nodes and $\mathcal{E}$ is the set of edges representing interactions. Each edge $(d_i, d_j, t) \in \mathcal{E}$ connects drugs $d_i$ and $d_j$ with a specific interaction type $t \in \mathcal{T}$.

The VGAE architecture consists of a graph-based encoder and a link prediction decoder.

\subsubsection{Graph Encoder}
The encoder is a Graph Neural Network (GNN), such as a Relational Graph Convolutional Network (R-GCN), that learns a latent representation for each drug. It takes the entire drug graph as input and produces a matrix of latent vectors $\mathbf{Z} \in \mathbb{R}^{|\mathcal{D}| \times K}$, where each row $\mathbf{z}_d$ is the embedding for drug $d$. The encoder, parameterized by $\phi$, defines a variational posterior $q_\phi(\mathbf{Z} | \mathcal{G})$ that approximates the true posterior. We assume a factorized Gaussian distribution:
$$q_\phi(\mathbf{Z} | \mathcal{G}) = \prod_{i=1}^{|\mathcal{D}|} q_\phi(\mathbf{z}_i | \mathcal{G}) = \prod_{i=1}^{|\mathcal{D}|} \mathcal{N}(\mathbf{z}_i | \boldsymbol{\mu}_i, \text{diag}(\boldsymbol{\sigma}_i^2))$$
where the means $\boldsymbol{\mu}$ and variances $\boldsymbol{\sigma}^2$ are the outputs of the GNN.

\subsubsection{Link Prediction Decoder}
The decoder, parameterized by $\theta$, reconstructs the DDI graph from the latent embeddings. For any pair of drugs $(d_i, d_j)$, it predicts the probability of each interaction type $t \in \mathcal{T}$. We use a multi-relational decoder, such as a DistMult model or a simple Multi-Layer Perceptron (MLP):
$$p_\theta(t | \mathbf{z}_i, \mathbf{z}_j) = \frac{\exp(f_\theta(\mathbf{z}_i, \mathbf{z}_j, t))}{\sum_{t' \in \mathcal{T}} \exp(f_\theta(\mathbf{z}_i, \mathbf{z}_j, t'))}$$
where $f_\theta$ is a scoring function (e.g., $\mathbf{z}_i^\top \mathbf{R}_t \mathbf{z}_j$ for DistMult, where $\mathbf{R}_t$ is a diagonal matrix for type $t$).

\subsubsection{VGAE Objective}
The VGAE is trained by maximizing the evidence lower bound (ELBO) on the training data $\mathcal{D}_{\text{train}}$:
$$\mathcal{L}_{\text{VGAE}}(\theta, \phi) = \mathbb{E}_{q_\phi(\mathbf{Z} | \mathcal{G}_{\text{train}})} \left[ \sum_{(d_i, d_j, t) \in \mathcal{E}_{\text{train}}} \log p_\theta(t | \mathbf{z}_i, \mathbf{z}_j) \right] - \text{KL}\left(q_\phi(\mathbf{Z} | \mathcal{G}_{\text{train}}) \,\|\, p(\mathbf{Z})\right)$$
where $p(\mathbf{Z}) = \prod_i \mathcal{N}(\mathbf{z}_i | 0, \mathbf{I})$ is the standard Gaussian prior and KL is the Kullback-Leibler divergence.

\subsection{GFlowNet for Balanced Synthetic DDI Generation}
The core issue with the initial VGAE is that it will be biased towards frequent interaction types. To mitigate this, we employ a GFlowNet to generate a supplementary dataset of synthetic DDIs, with a preference for rare types.

A GFlowNet learns a policy to construct an object $x$ through a sequence of actions, such that the probability of generating $x$ is proportional to a given reward $R(x)$.

\subsubsection{Trajectory and State Space}
We define the generation of a DDI triple $x = (d_i, d_j, t)$ as a three-step trajectory:
\begin{enumerate}
    \item \textbf{Initial State ($s_0$)}: The empty set.
    \item \textbf{Action 1 (Select Type)}: From $s_0$, select an interaction type $t \in \mathcal{T}$. The new state is $s_1 = (t)$.
    \item \textbf{Action 2 (Select First Drug)}: From $s_1$, select the first drug $d_i \in \mathcal{D}$. The new state is $s_2 = (t, d_i)$.
    \item \textbf{Action 3 (Select Second Drug)}: From $s_2$, select the second drug $d_j \in \mathcal{D} \setminus \{d_i\}$. The final state is the terminal state $s_f = (t, d_i, d_j)$.
\end{enumerate}

\subsubsection{Reward Function}
The reward function $R(x)$ guides the GFlowNet to generate valuable samples. We design a composite reward that balances \textbf{rareness} and \textbf{plausibility}:
$$R(t, d_i, d_j) = \underbrace{\left(\frac{1}{n_t + 1}\right)^\alpha}_{\text{Rareness}} \times \underbrace{p_\theta(t | \mathbf{z}_i, \mathbf{z}_j)}_{\text{Plausibility}}$$
\begin{enumerate}
    \item \textbf{Rareness}: $n_t$ is the frequency of interaction type $t$ in $\mathcal{D}_{\text{train}}$. The hyperparameter $\alpha \ge 0$ controls the strength of the emphasis on rare classes.
    \item \textbf{Plausibility}: $p_\theta(t | \mathbf{z}_i, \mathbf{z}_j)$ is the probability assigned by the pre-trained VGAE decoder. This ensures that the generated samples are consistent with the learned embedding space.
\end{enumerate}

\begin{figure}
    \centering
    \includegraphics[width=\linewidth]{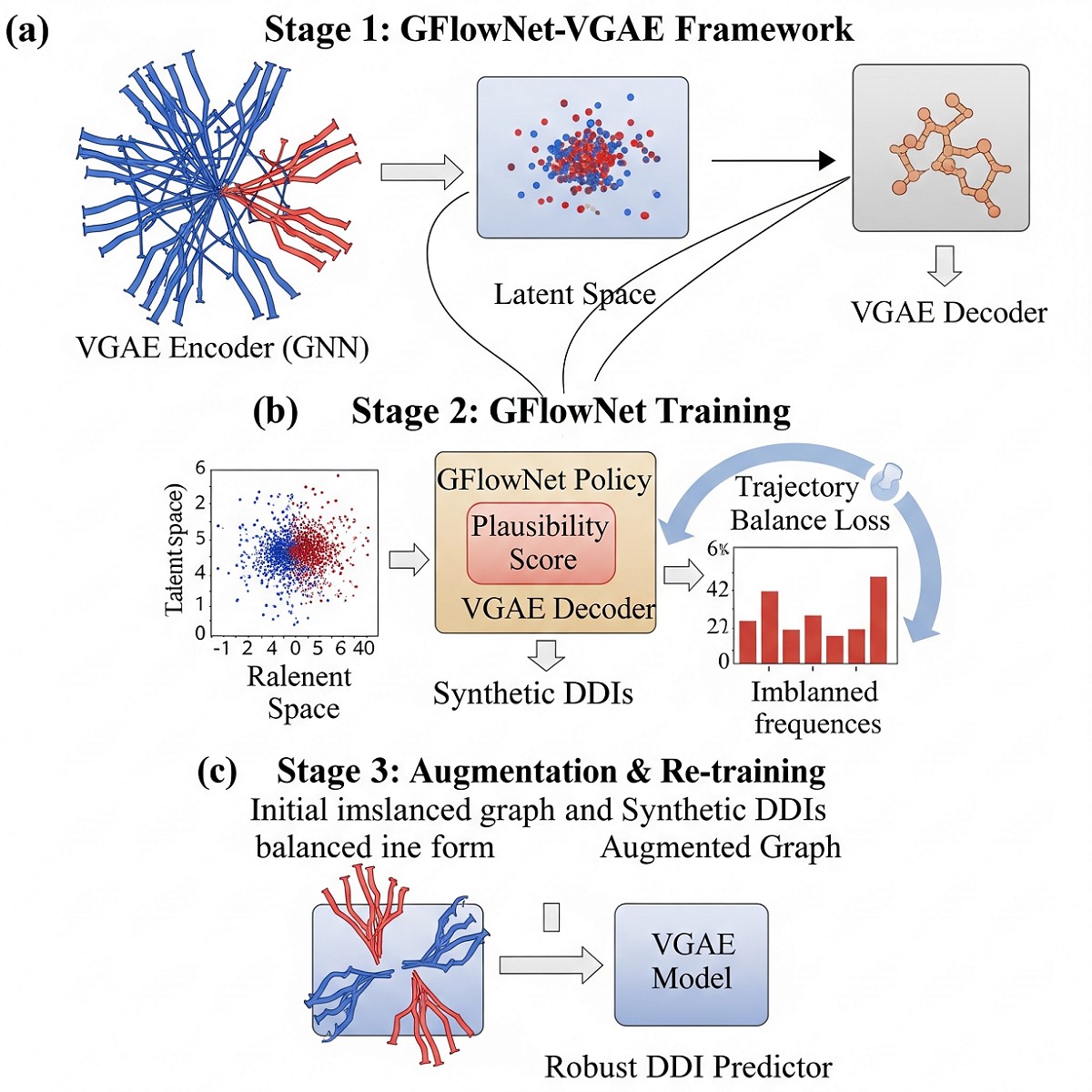}
    \caption{The proposed end-to-end framework. (a) A VGAE is first pre-trained on the original imbalanced DDI graph to learn drug embeddings. (b) A GFlowNet is then trained, using a reward function that combines plausibility from the VGAE and a rareness score, to learn a policy for generating synthetic DDIs. (c) The original data is augmented with the GFlowNet samples and used to train the final, robust VGAE model.}
    \label{fig:epipeline}
\end{figure}

\subsubsection{GFlowNet Policy and Training}
The GFlowNet learns a stochastic forward policy $P_F(s'|s; \psi)$, parameterized by a neural network $\psi$, for transitioning between states. To make training feasible, for Action 3 (selecting $d_j$), we restrict the action space. Given state $s_2=(t, d_i)$, the candidate set for $d_j$ is constructed from the $K$-nearest neighbors of $d_i$ in the VGAE's latent space $\mathbf{Z}$.

We train the policy network $\psi$ using the **Trajectory Balance (TB)** loss. The TB loss enforces a flow-matching condition over complete trajectories $\tau = (s_0 \to s_1 \to \dots \to s_f)$:
$$\mathcal{L}_{\text{TB}}(\psi) = \left( \log \frac{Z_\psi \prod_{s \to s' \in \tau} P_F(s'|s; \psi)}{R(s_f)} \right)^2$$
Here, $Z_\psi$ is a learnable parameter representing the total flow (partition function). Minimizing this loss encourages the learned sampling distribution $P_\text{GFN}(x)$ to be proportional to the reward $R(x)$.

\begin{figure}
    \centering
    \includegraphics[width=\linewidth]{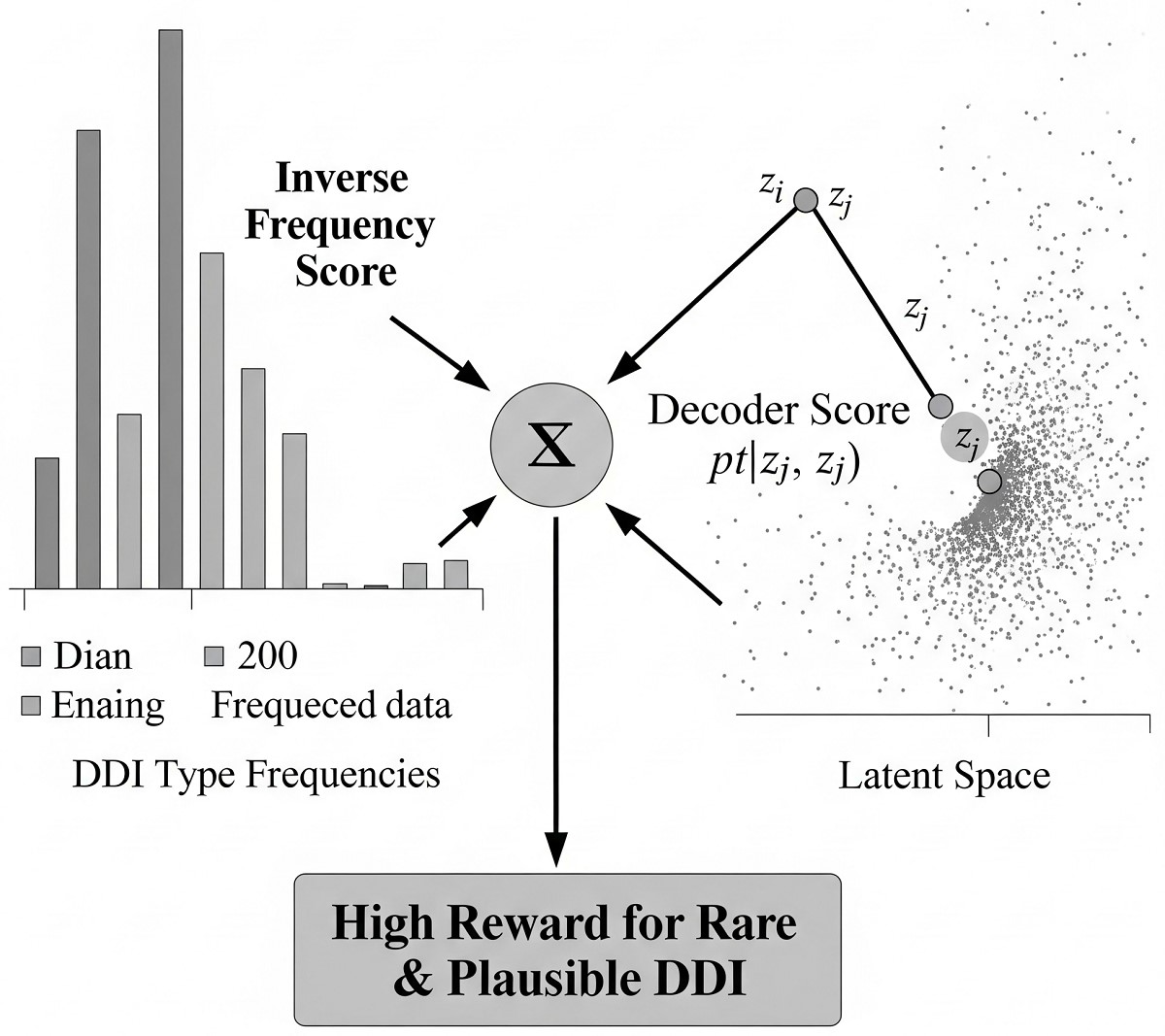}
    \caption{Composition of the GFlowNet reward function. The reward for a generated sample $(d_i, d_j, t)$ is the product of a rareness score, which is inversely proportional to the type's frequency in the training data, and a plausibility score, derived from the pre-trained VGAE decoder's confidence.}
    \label{fig:enter-label}
\end{figure}

\subsection{End-to-End Training Pipeline}
The complete training process involves three main stages:
\begin{enumerate}
    \item \textbf{Stage 1: VGAE Pre-training}: The VGAE model $(\theta, \phi)$ is trained on the original, imbalanced dataset $\mathcal{D}_{\text{train}}$ by minimizing $\mathcal{L}_{\text{VGAE}}$. This yields initial drug embeddings $\mathbf{Z}$ and a plausible decoder $p_\theta$.
    \item \textbf{Stage 2: GFlowNet Training}: Using the pre-trained embeddings $\mathbf{Z}$ and decoder $p_\theta$ to compute the reward, the GFlowNet policy $P_F(\cdot|\cdot; \psi)$ is trained by minimizing the Trajectory Balance loss $\mathcal{L}_{\text{TB}}$.
    \item \textbf{Stage 3: Augmentation and Final VGAE Training}: The trained GFlowNet policy is used to sample a set of $N$ synthetic DDIs, $\mathcal{D}_{\text{synth}}$. The original VGAE model is then re-trained (or fine-tuned) on the augmented dataset $\mathcal{D}_{\text{aug}} = \mathcal{D}_{\text{train}} \cup \mathcal{D}_{\text{synth}}$, again by minimizing $\mathcal{L}_{\text{VGAE}}$. This final model is used for DDI prediction.
\end{enumerate}

\begin{algorithm}
\caption{End-to-End Training of GFlowNet-VGAE for DDI Prediction}
\label{alg:main}
\begin{algorithmic}[1]
\State \textbf{Input:} DDI graph $\mathcal{G}_{\text{train}}$, number of synthetic samples $N$, GFN candidate size $K$.
\State \textbf{Parameters:} VGAE parameters $(\theta, \phi)$, GFlowNet policy parameters $\psi$.

\Function{TrainVGAE}{$\mathcal{G}$, epochs}
    \State Initialize VGAE parameters $(\theta, \phi)$.
    \For{epoch = 1 to epochs}
        \State Sample a batch of edges from $\mathcal{G}$.
        \State Compute latent embeddings $\mathbf{Z} \sim q_\phi(\cdot|\mathcal{G})$.
        \State Calculate $\mathcal{L}_{\text{VGAE}}$ using the batch.
        \State Update $(\theta, \phi)$ via gradient descent on $\mathcal{L}_{\text{VGAE}}$.
    \EndFor
    \State \textbf{return} Trained $(\theta, \phi)$ and final embeddings $\mathbf{Z}$.
\EndFunction

\Function{TrainGFlowNet}{$\mathcal{G}_{\text{train}}$, $(\theta, \phi)$, $\mathbf{Z}$, epochs}
    \State Initialize GFlowNet policy parameters $\psi$ and total flow $Z_\psi$.
    \For{epoch = 1 to epochs}
        \State \textit{// Sample a trajectory $\tau$}
        \State $s_0 \leftarrow \emptyset$
        \State $t \sim P_F(\cdot|s_0; \psi)$ \Comment{Sample interaction type}
        \State $s_1 \leftarrow (t)$
        \State $d_i \sim P_F(\cdot|s_1; \psi)$ \Comment{Sample first drug}
        \State $s_2 \leftarrow (t, d_i)$
        \State Construct candidate set $\mathcal{C}_{d_i}$ using $K$-NN of $d_i$ in $\mathbf{Z}$.
        \State $d_j \sim P_F(\cdot|s_2, \mathcal{C}_{d_i}; \psi)$ \Comment{Sample second drug from candidates}
        \State $s_f \leftarrow (t, d_i, d_j)$

        \State \textit{// Compute reward and loss}
        \State Calculate reward $R(s_f)$ using $p_\theta(t|\mathbf{z}_i, \mathbf{z}_j)$ and frequency of $t$.
        \State Calculate trajectory log-probability $\log P(\tau) = \sum_{s \to s' \in \tau} \log P_F(s'|s)$.
        \State Calculate $\mathcal{L}_{\text{TB}} = (\log Z_\psi + \log P(\tau) - \log R(s_f))^2$.
        \State Update $\psi$ and $Z_\psi$ via gradient descent on $\mathcal{L}_{\text{TB}}$.
    \EndFor
    \State \textbf{return} Trained policy network $\psi$.
\EndFunction

\State \textit{// Stage 1: Pre-train VGAE}
\State $(\theta_{\text{pre}}, \phi_{\text{pre}}), \mathbf{Z}_{\text{pre}} \leftarrow \text{TrainVGAE}(\mathcal{G}_{\text{train}}, \text{epochs}_1)$

\State \textit{// Stage 2: Train GFlowNet}
\State $\psi_{\text{trained}} \leftarrow \text{TrainGFlowNet}(\mathcal{G}_{\text{train}}, (\theta_{\text{pre}}, \phi_{\text{pre}}), \mathbf{Z}_{\text{pre}}, \text{epochs}_2)$

\State \textit{// Stage 3: Augment and Re-train}
\State $\mathcal{D}_{\text{synth}} \leftarrow \emptyset$
\For{$i = 1$ to $N$}
    \State Sample a synthetic DDI $(d_i, d_j, t)$ using the trained policy $\psi_{\text{trained}}$.
    \State $\mathcal{D}_{\text{synth}} \leftarrow \mathcal{D}_{\text{synth}} \cup \{(d_i, d_j, t)\}$.
\EndFor
\State $\mathcal{G}_{\text{aug}} \leftarrow \mathcal{G}_{\text{train}} \cup \text{edges from } \mathcal{D}_{\text{synth}}$.
\State $(\theta_{\text{final}}, \phi_{\text{final}}), \mathbf{Z}_{\text{final}} \leftarrow \text{TrainVGAE}(\mathcal{G}_{\text{aug}}, \text{epochs}_3)$ \Comment{Re-train on augmented data}

\State \textbf{Output:} The final, robust DDI prediction model $(\theta_{\text{final}}, \phi_{\text{final}})$.
\end{algorithmic}
\end{algorithm}

\section{Experiments}
\subsection{Dataset}
In this preliminary work, we use the DrugBank dataset \cite{Wishart2018-ae}, which includes 1,703 drugs and 191,870 drug pairs spanning 86 DDI types, along with structural and chemical information. The dataset was split into three subsets: 115,185 drug pairs for training, 38,348 for validation, and 38,337 for testing.

\subsection{Implementation Details}
We implemented VGAE for DDI prediction  \cite{ngo2022predictingdrugdruginteractionsusing}, given its generative nature. The results in Table \ref{tab:res} show the impact of adding the GFlowNet module to the VGAE framework for DDI prediction. AUROC, Accuracy, AUPRC, and F1 scores were all above 0.99, with minor differences in classification metrics, suggesting GFlowNet had little impact on performance. However, diversity and coverage metrics showed substantial improvements, demonstrating GFlowNet’s effectiveness in addressing class imbalance.

\section{Results and Discussion}
To evaluate the effectiveness of our proposed generative augmentation strategy in addressing class imbalance in drug–drug interaction (DDI) prediction, we employ two key diversity metrics: \textit{Shannon Entropy (SE)} and \textit{Jensen–Shannon Divergence (JSD)}. Table \ref{tab:res} demonstrates the results of our experimentation. These metrics provide quantitative insights into the distributional properties of the interaction types before and after augmentation, beyond what is captured by standard classification metrics.

\textit{Shannon Entropy (SE)} \cite{Fang2008}, measures the uncertainty or diversity of a probability distribution. For a discrete distribution \( P = \{p_1, p_2, ..., p_n\} \), the Shannon Entropy is computed as:

\begin{equation}
H(P) = -\sum_{i=1}^{n} p_i \log_2 p_i,
\end{equation}

where \( p_i \) denotes the probability of the \( i \)-th interaction type. A higher entropy value indicates a more uniform distribution across interaction types, signifying reduced class imbalance. In our experiments, the SE increased from 1.23 to 1.69 after applying the GFlowNet-based augmentation. This reflects a more balanced representation of interaction types in the dataset, which is especially beneficial for improving the learning of under-represented classes.

\textit{Jensen–Shannon Divergence (JSD)}\cite{Menndez1997}, quantifies the similarity between two probability distributions. For two distributions \( P \) (augmented) and \( Q \) (true), JSD is defined as:

\begin{equation}
JSD(P || Q) = \frac{1}{2} D_{KL}(P || M) + \frac{1}{2} D_{KL}(Q || M),
\end{equation}

where \( M = \frac{1}{2}(P + Q) \) is the average of the two distributions, and \( D_{KL} \) denotes the Kullback–Leibler divergence. JSD is symmetric and bounded between 0 and 1, where lower values indicate greater similarity. In our case, JSD decreased from 0.35 to 0.12, indicating that the distribution of synthetic samples more closely matches the empirical distribution of real interactions. This alignment ensures that the generative process is not just balancing classes arbitrarily but doing so in a way that reflects the true underlying data distribution.

Together, the increase in Shannon Entropy and the decrease in Jensen–Shannon Divergence demonstrate that our framework successfully promotes a more balanced and representative dataset. This, in turn, facilitates improved generalization, particularly for rare interaction types that are often clinically critical. The observed improvements in coverage—from 0.2441 to 0.7709—further support the model's enhanced capacity to capture and learn from these rare interactions. While traditional metrics showed limited change, these diversity-aware evaluations highlight the strength of generative modeling in addressing real-world challenges in biomedical prediction tasks.

\begin{table}[t]
    \centering
    \caption{Experimental Findings}
    \label{tab:res}
    \begin{tabular}{cccccccc}
    \hline
       Setup & AUROC  & Accuracy & AUPRC & F1 Score & SE & JSV & Coverage\\ \hline
        Without GFN & 0.99081 & 0.96859 & 0.98861 & 0.98982 & 1.23 & 0.35 & 0.2441 \\
        With GFN & 0.99071 & 0.96792 & 0.98922 & 0.99914 & 1.69 & 0.12 & 0.7709 \\
        \hline
    \end{tabular}
\end{table}

\section{Discussion}
Our proposed framework addresses a longstanding challenge in drug–drug interaction (DDI) prediction: the pervasive class imbalance that limits model generalization and risks overlooking rare yet clinically significant interactions. By integrating a GFlowNet with a Variational Graph Autoencoder (VGAE), we introduce a principled mechanism for generating synthetic DDI samples conditioned on interaction types. The reward-driven sampling process encourages the generation of under-represented classes, leading to improved balance in the training data, as evidenced by higher Shannon Entropy and lower Jensen–Shannon Divergence. This shift enhances the model's ability to learn from both frequent and rare interactions, which is crucial in clinical contexts where even infrequent interactions can have serious consequences. Our improvements in diversity metrics and coverage suggest that generative methods, when guided by structural priors and task-specific incentives, offer a promising alternative to traditional oversampling or class-weighting strategies.

Beyond immediate performance gains, this work lays the foundation for broader applications in imbalanced biomedical prediction tasks, where generative augmentation can act as a form of informed exploration. By aligning synthetic distributions with real-world priors, our approach supports more equitable representation of minority classes—an issue that extends across genomics, adverse event prediction, and rare disease modeling. Furthermore, the modular design of our framework allows for easy extension: future work could explore integrating domain-specific constraints, leveraging external knowledge graphs, or employing reward shaping to target high-risk interaction types. Ultimately, our contribution not only advances the technical state-of-the-art but also pushes the field toward more robust and fair AI systems that are better aligned with the needs of healthcare and biomedical decision-making.

\section{Conclusion}
This work introduces a novel approach to tackling class imbalance in drug–drug interaction prediction by integrating a GFlowNet-based generative sampling strategy into a VGAE framework. While traditional classification metrics remain largely unchanged, diversity-aware evaluations show significant improvements, highlighting GFlowNet’s ability to discover under-represented interaction types, crucial for patient safety in clinical applications.
Future work will focus on extending the framework to larger, more diverse datasets and optimizing sampling strategies to improve rare class representation.

\bibliographystyle{splncs04}
\bibliography{mybibliography}

\end{document}